\documentclass[10pt,twocolumn,letterpaper]{article}

\usepackage{cvpr}
\usepackage{times}
\usepackage{epsfig}
\usepackage{graphicx}
\usepackage{amsmath}
\usepackage{amssymb}

\usepackage{epsfig}
\usepackage{subfigure}
\usepackage{graphicx}
\usepackage{algorithm}
\usepackage{algorithmic}
\usepackage{amssymb}
\usepackage{mathrsfs} 
\usepackage{amsmath,bm}
\usepackage{array,multirow}
\usepackage{mdwlist}
\usepackage{paralist}
\usepackage{url}
\usepackage{color}
\usepackage{caption}
\usepackage{setspace}
\usepackage{textcomp}
\usepackage{bbding}
\usepackage{booktabs}
\usepackage{threeparttable}

\usepackage[pagebackref=true,breaklinks=true,letterpaper=true,colorlinks,bookmarks=false]{hyperref}
\graphicspath{{figures/}}

\cvprfinalcopy 

\def\cvprPaperID{XXXX} 

\ifcvprfinal\fi
\begin{document}

\title{Single-Shot Bidirectional Pyramid Networks for High-Quality Object Detection} 

\author{Xiongwei Wu$^\dagger$, \quad  Daoxin Zhang$^\dagger$$^\ddagger$, \quad Jianke Zhu$^\ddagger$,\quad Steven C.H. Hoi$^\dagger$$^\S$\\
$^\dagger$School of Information Systems, Singapore Management University, Singapore\\
$^\ddagger$College of Computer Science and Technology, Zhejiang University, Hangzhou, China\\
$^\S$DeepIR Inc., Beijing, China\\
{\tt\small \{chhoi,dxzhang,xwwu.2015@phdis\}@smu.edu.sg};{\tt\small\{dxzhang,jkzhu\}@zju.edu.cn}\\
}

\maketitle

\begin{abstract}
Recent years have witnessed many exciting achievements for object detection using
deep learning techniques. Despite achieving significant progresses, most existing
detectors are designed to detect objects with relatively low-quality prediction of
locations, i.e., often trained with the threshold of Intersection over Union (IoU) set to
0.5 by default, which can yield low-quality or even noisy detections. It remains an open
challenge for how to devise and train a high-quality detector that can achieve more
precise localization (i.e., IoU$>$0.5) without sacrificing the detection performance. In
this paper, we propose a novel single-shot detection framework of Bidirectional
Pyramid Networks (BPN) towards high-quality object detection, which consists of two
novel components: (i) a Bidirectional Feature Pyramid structure for more effective and
robust feature representations; and (ii) a Cascade Anchor Refinement to gradually
refine the quality of predesigned anchors for more effective training. Our experiments
showed that the proposed BPN achieves the best performances among all the
single-stage object detectors on both PASCAL VOC and MS COCO datasets, especially for
high-quality detections.
\end{abstract}

\section{Introduction}

Object detection is one of fundamental research problems in computer vision and has
been extensively studied in literature\cite{ren2015faster,girshick2014rich,lin2016fpn}.
Recent years have witnessed remarkable progresses for object detection after exploring
the family of powerful deep learning techniques. Currently, the state-of-the-art deep
learning based object detection frameworks can be generally divided into two major
groups: (i) two-stage detectors, such as the family of Region-based CNN (R-CNN) and
their variants\cite{girshick2015fast,ren2015faster,girshick2014rich} and (ii) one-stage
detectors, such as SSD and its variants\cite{redmon2016yolo9000,liu2016ssd}. Two-stage
RCNN-based detectors first learn to generate a sparse set of proposals followed by
training region classifiers, while one-stage SSD-like detectors directly make categorical
prediction of objects based on the predefined anchors on the feature maps without the
proposal generation step. Two-stage detectors usually achieve better detection
performance and often report state-of-the-art results on benchmark data sets, while
one-stage detectors are significantly more efficient and thus more suitable for many
real-word practical/industrial applications where fast/real-time detection speed is of
crucial importance.

Despite being studied extensively, most existing object detectors are designed for
achieving localization with relatively low-quality precision, i.e., with a default IoU
threshold of 0.5. When the goal is to achieve higher quality localization precision
(IoU$>$0.5), the detection performance often drops significantly\cite{cai2017cascade}.
One naive solution is to increase the IoU threshold during training. This however is not
effective since a high IoU will lead to significantly less amount of positive training
samples and thus make the training results prone to overfitting, especially for single-shot
SSD-like detectors. This work is motivated to investigate an effective single-shot
detection scheme towards high-quality object detection.

In this paper, we aim to develop a novel high-quality detector by following the family of
single-stage SSD-like detectors due to their significant advantage in computational
efficiency. In particular, we realize that the existing SSD-style detector has two critical
drawbacks for high-quality object detection tasks. First, the single-shot feature
representations may not be discriminative and robust enough for precise localization.
Second, the singe-stage detection scheme relies on the predefined anchors which are
very rigid and often inaccurate. To overcome these drawbacks for high-quality object
detection tasks, in this paper, we propose a novel single-shot detection framework
named ``Bidirectional Pyramid Networks" (BPN). As a summary, our main contributions
include the following:
\begin{itemize}
\item A novel framework of Bidirectional Pyramid Networks (BPN) for single-shot object
    detector that is designed directly towards high-quality detection;
\item A novel Bidirection Feature Pyramid Structure that improves the vanilla Feature
    Pyramid by adding a Reverse Feature Pyramid in order to fuse both deep and shadow
    features towards more effective and robust representations;
\item A novel Cascaded Anchor Refinement scheme to gradually improve the quality of
    predefined anchors which are often inaccurate at the beginning;
\item Extensive experiments on PASCAL VOC and MSCOCO showed that the proposed
    method achieved the state-of-the-art results for high-quality object detection while
    maintaining the advantage of computational efficiency.
\end{itemize}

\section{Related Work}
Object detection has been extensively studied for decades
\cite{viola2004robust,girshick2014rich,felzenszwalb2010object}. In early stage of
research studies, object detection was based on sliding windows, and dense image grids
are encoded with hand-crafted features followed by training classifiers to find and locate
objects. Viola and Jones \cite{viola2004robust} proposed a pioneering cascaded
classifiers by AdaBoost with Haar feature for face detection and obtained excellent
performance with high efficiency. After the remarkable success of applying Deep
Convolutional Neural Networks on image classification tasks
\cite{he2016deep,simonyan2014very,krizhevsky2012imagenet}, deep learning based
approaches have been actively explored for object detection, especially for the
region-based convolutional neural networks (R-CNN) and its variants
\cite{ren2015faster,lin2016fpn,girshick2015fast}. Currently deep learning based
detectors can be generally divided into two groups: (i) two-stage RCNN-based methods
and (ii) one-stage SSD-based methods. RCNN-based methods, such as RCNN
\cite{girshick2014rich}, Fast RCNN \cite{girshick2015fast}, Faster RCNN
\cite{ren2015faster}, R-FCN \cite{dai2016r}, first generate a sparse set of proposals
followed by region classifiers and location regressors. Two-stage detectors usually
achieve better detection performance and report state-of-the-art results on many
common benchmarks because the proposals are often carefully generated (e.g., by
selective search \cite{uijlings2013selective} or RPN \cite{ren2015faster}) and usually well
match the target objects. However, they often suffer from very slow inference speed due
to the expensive two-stage detection approach. Unlike the two-stage RCNN-based
methods, SSD-style methods, such as SSD \cite{liu2016ssd}, YOLO \cite{redmon2016you},
YOLOv2 \cite{redmon2016yolo9000}), ignores the proposal generation step by directly
making prediction with manually pre-defined anchors and thus reduce the inference
time significantly towards real-time speed. However, the manually predefined anchors
are often sub-optimal and sometimes ill-designed, in which few of them can tightly
match the objects. Thus, SSD-style detectors \cite{liu2016ssd} are difficult to precisely
locate objects towards high-quality detection.

In literature, most object detection studies were focused on the detection with relatively
low localization quality, with a default IoU threshold of 0.5. There are only a few related
studies for high-quality detection. LocNet\cite{gidaris2016locnet} learns a
postprocessing network for location refinement, which however does not optimize the
whole system end-to-end and is not designed for high-quality detection tasks. MultiPath
Network \cite{zagoruyko2016multipath} proposed to learn multiple detection branches
for different quality thresholds. However, it still suffers insufficient training samples and
it is computationally slow due to the nature of two-stage detectors. Cascaded RCNN
\cite{cai2017cascade} learns regressors in a cascaded way, which gradually increases
qualified proposal numbers towards high-quality detection. However, it is still based on
two-stage RCNN and its slow inference speed is a critical drawback, especially the
feature re-extraction step is operated by time-consuming ROI Pooling or ROI Warping.

Our work is also related to the studies for multi-scale feature fusion, which has been
proved to be an effective and important structure for object detection with different
scales. ION \cite{bell2016inside} extracts region features from different layers by ROI
Pooling operation; HyperNet \cite{kong2016hypernet} directly concatenates features at
different layers using deconvolution layers. FPN \cite{lin2016fpn} and
DSSD\cite{fu2017dssd} fuses features of different scales with lateral connection in a
bottom-up manneer, which effectively improve the detection of small objects. However,
the vanilla feature pyramid \cite{lin2016fpn} only considers boosting shallow layer
features with deep layer features, but ignores the fact that the instance information in
shallow layer features can be helpful to deep semantic layer features. We overcome this
limitation by the proposed Bidirectional Feature Pyramid structure.

\section{Single-Shot Detector for High-Quality Detection}
\subsection{Motivation}

Our goal is to investigate single-shot detectors for high-quality object detection tasks.
For existing object detectors, a group of anchors are often generated/pre-defined on the
feature maps densely or sparsely, followed by location regression and object category
classification. The object class label of each anchor is assigned according to anchor's
jaccard overlap with objects. Two-stage RCNN-based detectors generate anchors in their
first step, and assign positive label to anchors whose overlaps with objects are higher
than IoU threshold. However, in one-stage SSD-based detectors, anchors are manually
designed and thus the majority of these anchors fail to match objects with qualified IoU
threshold(0.5 etc.). This problem becomes more severe in training detectors for high IoU
thresholds(0.7 etc.) since the number of positive anchors decreases significantly as IoU
thresholds increase, and it leads to overfitting problem.

In the first column of Table \ref{tab:matchnum}, we count the number of positive
anchors per image for different IoU thresholds in training SSD-style detector. In default
setting(IoU threshold is 0.5), the positive anchor number per image is only 13.85. When
we increase IoU threshold from 0.5 to 0.7, only 3 positive anchors left for training, which
cannot provide sufficient information to effectively train detectors. Our motivation is to
improve anchor quality by cascaded refine pre-designed anchor in different predict
levels. The second and third column in Table \ref{tab:matchnum} illustrate the matched
anchor number after once and twice being refined respectively. With sufficient matched
anchors, we are able to train high quality detector in Bi-directional feature pyramid.
Please refer Section \ref{subsec:abla} for details.

\begin{table}[t]
\centering
\caption{Average matched anchor number per image whose jaccard overlaps
with objects are higher than IoU Threshold. Our cascaded anchor refiner improves anchor
quality gradually. Underlined entry is the number we use for training(discussed in section\ref{subsec:abla}). }
\scriptsize
\begin{tabular}{c|c|c|c}
\toprule[1.5pt]
\multirow{2}{*}{IoU Threshold} &\multirow{2}{*}{Original SSD} &\multirow{2}{*}{FPN(once refined)} &\multirow{2}{*}{BPN(twice refined)} \\
  & &  &\\
\hline
0.5 &\underline{\textbf{13.85}} &191.11&383.09\\
0.6 &5.08 &\underline{\textbf{141.68}}&316.86\\
0.7  &3.01 &100.52&\underline{\textbf{252.16}}\\
0.8  &2.84&62.12 &176.26\\
0.9  &2.84 &22.18&64.21\\
\bottomrule[1.5pt]
\end{tabular}
\label{tab:matchnum}
\end{table}

\begin{figure*}[htb]
\centering
\includegraphics[height=7cm]{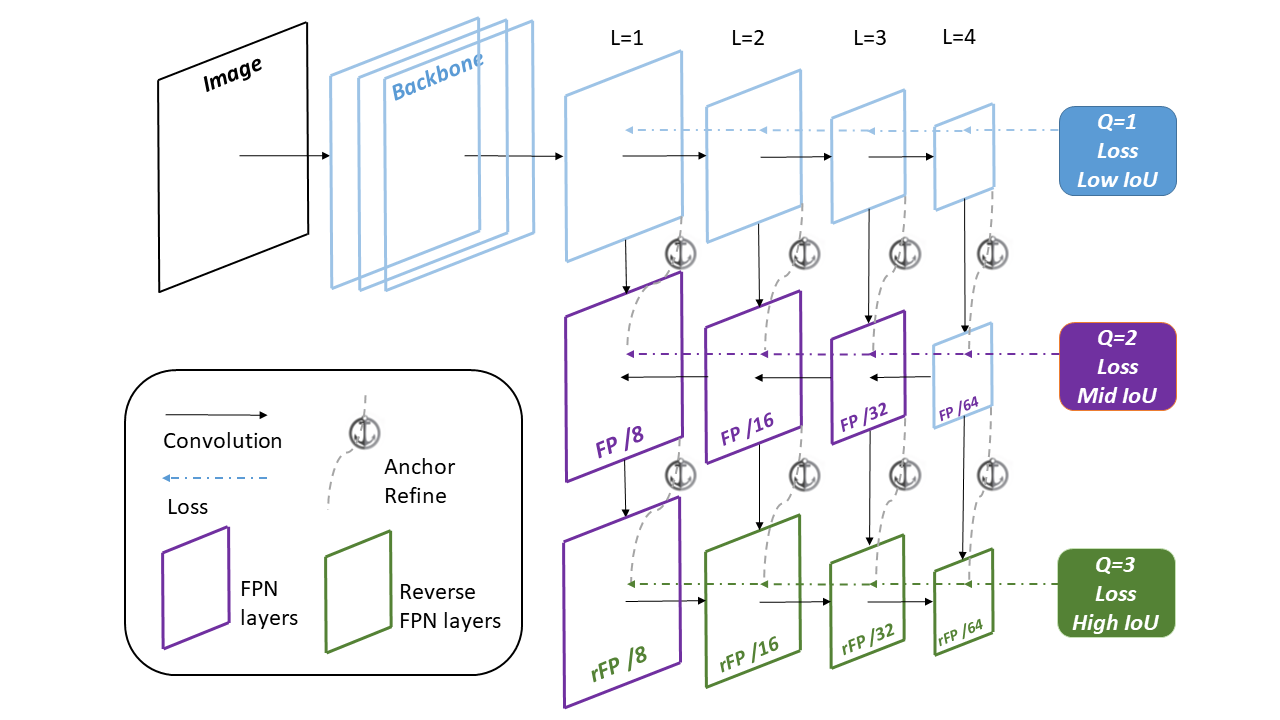}
\caption{The proposed framework of Bidirectional Pyramid Networks (BPN) for single-shot high-quality detection. The Cascaded Anchor Refinement (\textit{CAR}) are utilized for relocating anchors,
each of which is responsible for a certain quality level of detection. Training sample quality increases when anchor refinement cascades (with higher IoU). \textit{FPN} denotes Feature Pyramid block, and \textit{rFPN} denotes Reverse Feature Pyramid block.}
\label{fig:pipeline}
\end{figure*}

\subsection{Proposed Framework of Bidirectional Pyramid Networks}
In this paper, we propose a novel framework of Bidirectional Pyramid Networks (BPN) to
overcome the above drawbacks of SSD-style detectors towards high-quality detection
tasks. In particular, to address the weak feature representation issue of SSD-style
detectors, the idea is to explore the structure of Feature Pyramid Networks (FPN)
\cite{lin2016fpn} in improving the typical SSD-style feature representations, in which we
propose a novel Bidirectional Feature Pyramid structure that can further boost the
effectiveness of FPN structure. To address the anchor quality issue, the key idea and
challenge is to devise an effective yet efficient scheme for refining the quality of the
anchors before training the classifiers and regressors, in which we explore a cascade
learning and refinement approach without suffering the computational drawback of
two-stage detectors.

Figure \ref{fig:pipeline} gives an overview of the proposed single-shot Bidirectional
Pyramid Networks (BPN) for high-quality object detection, where the backbone network
can be any typical CNN network, such as Alexnet\cite{krizhevsky2012imagenet},
GoogleNet\cite{szegedy2015going}, VGG\cite{simonyan2014very},
ResNet\cite{he2016deep}, etc, as shown in the blue branch of Figure \ref{fig:pipeline}.
For simplicity, we choose VGG-16 as backbone network in our study.

Similar to typical single-shot detectors, at the lowest quality level with the default
IoU=0.5, the proposed BPN detector makes the prediction based on the predefined
anchors. Then, the features are further enhanced by the Bidirectional Feature Pyramid
which aggregates features from different depths. It consists of standard feature pyramids
in bottom-up (the purple branch of Figure \ref{fig:pipeline}) and reverse feature pyramid
in top-down (the green branch of Figure \ref{fig:pipeline}). These three-level branches
not only aggregate multi-level features to provide robust feature representations, but
also enable multi-quality training and cascaded Anchor refinement. For the joint training
with multiple quality levels, the Cascaded Anchor Refinement optimizes anchors from
the previous branch and send them to the next branch.

The above two key components, Bidirectional Feature Pyramid and Cascaded Anchor
Refinement, are nicely integrated in the framework and can be trained end-to-end to
achieve high-quality detections in a coherent and synergic manner. In the following, we
present each of these two components in detail.

\subsection{Cascaded Anchor Refinement}
We denote the depth of feature maps for prediction as $L$, where $L \in \left \{ 1, 2, 3,4
\right \}$ in our settings, and the levels of quality $Q \in \left \{1, 2, 3,\ldots\right\}$ with
the corresponding IoU thresholds as $\mathrm{IoU}(Q) \in \left \{ 0.5, 0.6, 0.7,...
\right\}$. The feature map in depth $L$ for quality $Q$ prediction is denoted as
$F_L^Q$, and anchors for training quality $Q$ detector in depth $L$ is denoted as
$A^Q_L$. Specifically for this work, we choose three types of detectors with different
quality levels: \textit{Low}, \textit{Mid} and \textit{High} with the corresponding IoU
threshold as 0.5, 0.6 and 0.7 respectively (See Figure \ref{fig:pipeline} for details).

In order to increase the number of positive anchors and improve their quality as well, we
denote the Cascaded Anchor Refinement (``CAR") used in quality $Q$, depth $L$ as
CAR$^Q_L$. In particular, CAR has two parts: location regressor Reg$^Q_L$ and
categorical classifiers Cls$^Q_L$. At each level of quality, regressors receive the
processed anchors from the previous level of quality for further optimization ($A^1_L$ is
the manually defined anchors):
\begin{equation}
A^Q_L = Reg^Q(A^{Q-1}_L; F^Q_L),\quad Q=2,3,\ldots, L=1,2,\ldots
\end{equation}
Categorical classifiers learn to predict categorical confidence scores and assign them to
these anchors:
\begin{equation}
C^Q_L = Cls^Q(F^Q_L),\quad Q=1,2,3\ldots, L=1,2,\ldots
\end{equation}
Therefore, the training loss at quality level $Q$ can be represented as:
\begin{equation}
L^Q = \frac{1}{N_{\text{Q}}}*\sum_{L} \sum_{i} \Big(L_{\text{Cls}}^Q(\{C^{Q}_{L_i}\}, \{l_{L_i}\}) + \lambda*L_{\text{Reg}}^Q( \{A^{Q}_{L_i}\}, \{g_{L_i}\})\Big)\label{3}
\end{equation}
where ${N_{\text{Q}}}$ is the positive sample number at quality level $Q$, $L_i$ is the
index of anchor in depth $L$ feature map within a mini-batch, $l_{L_i}$ is the ground
truth class label of anchor $L_i$, $g_{L_i}$ is the ground truth location and size of anchor
$L_i$, $\lambda$ is the balance weighting parameter which is simply set to $1$ in our
settings. $L_{\text{Cls}}^Q(.)$ is softmax loss function over multiple classes confidences
and $L_{\text{Reg}}^Q(.)$ is the Smooth L1-loss which is also used in \cite{liu2016ssd}.
The total training loss is the summation of losses at all the quality levels:
\begin{equation}
 L_{\text{BPN}} = \sum_{Q}  L^Q
\end{equation}

\subsection{Bidirectional Feature Pyramid Structure}

In order to improve the power of feature representation of SSD-style detectors, we apply
 Feature Pyramid Networks (FPN) \cite{lin2016fpn}, which exploits the
inherents multi-scale, pyramidal hierarchy of deep convolutional networks to construct
the representation of feature pyramids with marginal extra cost. Specifically, FPN fuse
deep semantically-strong features with shallow semantically-weak but instance-strong
features. However, we found aggregating  features in the reverse direction is of great
importance similarly. This results in the proposed Bidirectional Feature Pyramid
structure that consists of both FPN and reverse FPN to make the feature representation
considerably more effective and robust.

\begin{figure*}[ht!]
\vspace{-0.5in}
\centering
	\subfigure[Feature Pyramid] {
		\includegraphics[width=0.5\textwidth]{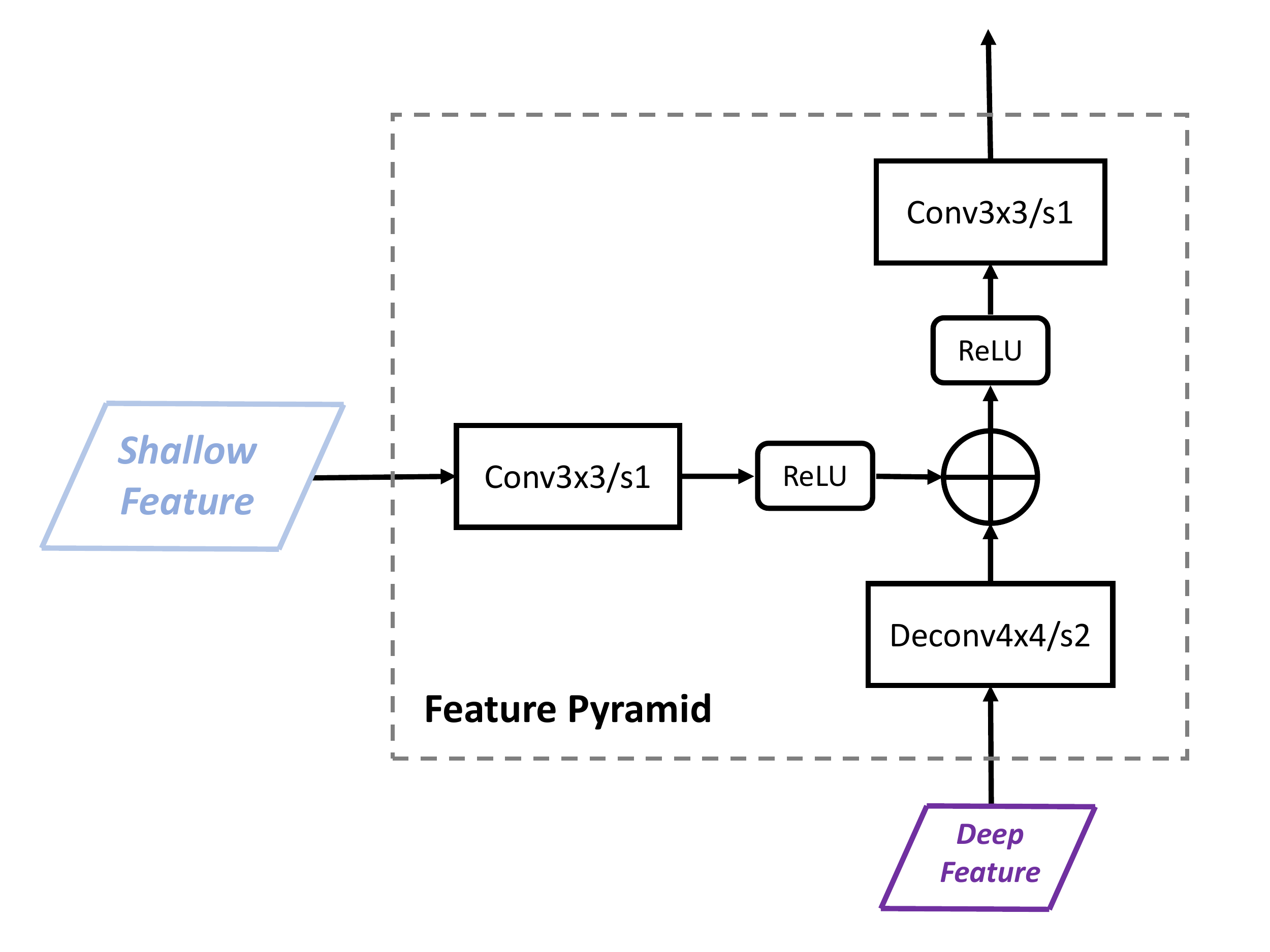}}
	\subfigure[Reverse Feature Pyramid]{
		\includegraphics[width=0.5\textwidth]{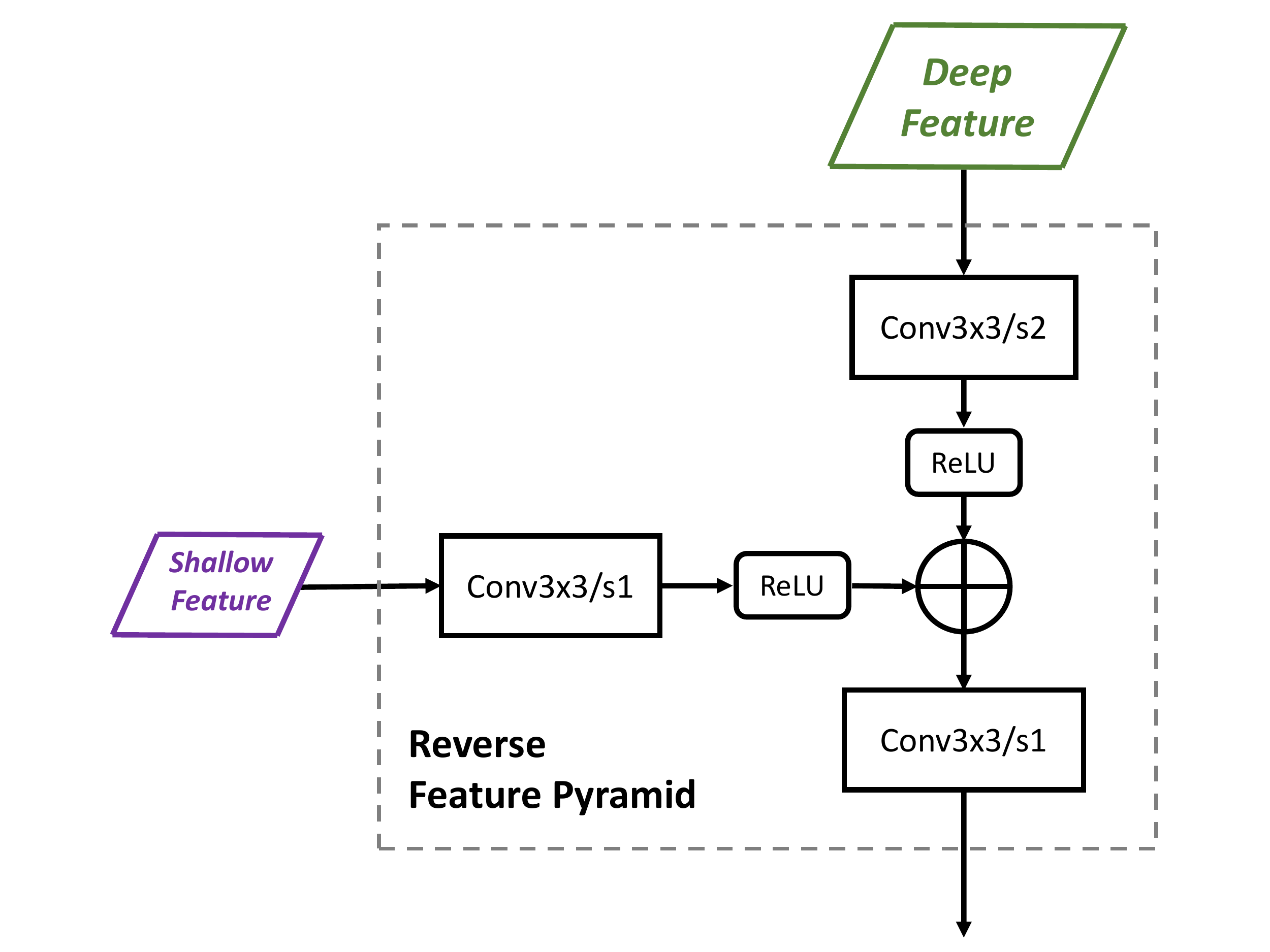}}
	\caption{The proposed Bidirectional Feature Pyramid Network Structure} \label{BPN}
\end{figure*}

Reverse FPN can enjoy the merit from following aspects: 1). Compared with stacked CNN
for image classification, reverse FPN reduces the \textit{distance} from shallow features
to deep features by using much fewer convolution filters and thus effectively keeps
spatial information; 2) Lateral connections \textit{reuse} different shallow layer features
to reduce information attenuation from shallow features to deep features; 3) Our
cascaded-style detector structure can naturally use this structure. Figure \ref{BPN} gives
the illustration of the proposed Bidirectional Feature Pyramid structure.

Specifically, Figure \ref{BPN}(a) is the vanilla Feature Pyramid block that fuses features in
bottom-up with lateral connection. It worth noticing that there is no strengthen of the
deepest feature layer from Feature Pyramid (the right diagram of
Figure\ref{fig:pipeline}). Thus, we further build the Reverse Feature Pyramid by
top-down aggregation (as shown in Figure \ref{BPN} (b)) with lateral connection to
enhance deep layer features. The formulation of Feature Pyramid (FP) and reverse
Feature Pyramid (rFP) can be represented as:
\begin{eqnarray}
\mathrm{FP:\quad} F_L^Q &=& \mathrm{Deconv}_{s2}(F_{L+1}^{Q}) \oplus \mathrm{Conv}(F_L^{Q-1})\\
\mathrm{rFP:\quad} F_L^Q &=& \mathrm{Conv}_{s2}(F_{L-1}^{Q}) \oplus \mathrm{Conv}(F_L^{Q-1})
\end{eqnarray}
where $\mathrm{Deconv}_{s2}$ denotes the deconvolution operation for feature map
up-sampling with stride 2 and $\mathrm{Conv}$ denotes convolution operation.
$\oplus$ denotes element-wise summation. In this paper, we use $3\times3$
convolution kernels with $256$ channels to build the Feature Pyramid and Reverse
Feature Pyramid.

\subsection{Implementation Details}
\noindent{\bf Backbone Architecture:} We choose VGG16 \cite{simonyan2014very}
pre-trained on ImageNet as the backbone network in our experiments. We follow
\cite{liu2016ssd} to transform the last two fully-connected layers ``fc6" and ``fc7" to
convolutional layers ``conv\_fc6" and  ``conv\_fc7" via reducing parameters. To increase
receptive fields and capture large objects, we attached two additional convolution layers
after the VGG16 (dabbed as conv6\_1 and conv6\_2). Due to different scale norm in
different feature maps, we re-scale the norms of the first two feature blocks to 10 and 8
respectively to stable the training process.

\noindent{\bf Data Augmentation:} We follow the data augmentation strategies in
\cite{liu2016ssd} to make the detectors robust to objects with multiple scales and colors.
Specifically, images are randomly expanded or cropped with additional photometric
distortion to generate more additional training samples.

\noindent{\bf Feature Blocks for Prediction:} In order to detect objects at different
scales, we use multiple feature maps for prediction. The vanilla convolution feature
blocks in VGG16 are used for low-quality detection, feature pyramid blocks are used for
mid-quality detection, and the reverse feature pyramid blocks are used for high-quality
detection. We use four feature blocks with stride 8, 16, 32 and 64 pixels in training each
quality detector (In VGG16, conv4\_3, conv5\_3,  conv\_fc7 and conv6\_2 and their
corresponding feature pyramid blocks FP3, FP4, FP5 and FP6, and reverse feature
pyramid blocks rFP3, rFP4, rFP5 and rFP6 are used etc.)

\noindent{\bf Anchor Design:} Originally a group of anchors are pre-designed manually.
For each prediction feature block, one scale-specific set of anchors with three aspect
ratios are associated. In our approach, we set the scale of anchors as 4 times of the
feature map stride and set the aspect ratios as 0.5, 1.0 and 2.0 to cover different scales of
objects. We first match each object to the anchor box with the best overlap score, and
then match the anchor boxes to any ground truth with overlap higher than the quality
thresholds.

\noindent{\bf Optimization Details:} We use the ``xavier" method in
\cite{glorot2010understanding} to randomly initialize the parameters in extra added
layers in VGG16. We set the  mini-batch size as 32 in training and the whole network is
optimized via the SGD optimizer. We set the momentum to 0.9 and weight decay to
0.005. The initial learning rate is set to 0.001. The learning policy is different for different
datasets. For PASCAL VOC dataset, the models are totally finetuned for 120k iterations
and we decrease the learning rate to 1e-4 and 1e-5 after 80k and 100k iterations,
respectively. For MSCOCO dataset, the models are finetuned for 400k iterations and we
decrease the learning rate to 1e-4 and 1e-5 after 280k and 360k iterations, respectively .
All the detectors are optimized end-to-end.

\noindent{\bf Sampling Strategy:} The ratio of positive and negative anchors are
imbalanced after anchor matching step, so proper sampling strategy is necessary to
stable training process. In this paper, we sample a subset of negative anchors to keep the
ratio of positive and negative anchors as 1:3 in training process. In order to stable
training and fast convergency, instead of randomly sampling negative anchors, we sort
the negative anchors according to their confidence loss values and select the hardest
ones for training. In different quality levels, the IoU thresholds for training are different.
In this paper, we set three quality levels: low, mid and high qualities with IoU thresholds
as 0.5, 0.6 and 0.7.

\noindent{\bf Inference:} During the inference phase, different quality CARs make
prediction and send the refined anchors into next quality level. We take the predictions
from CARs in all qualities, to make sure that it is suitable for all the low-, mid- and
high-quality detection.

\section{Experiments}
We conduct extensive experiments on two public benchmarks Pascal VOC and MSCOCO.
Pascal VOC has 20 categories and MSCOCO has 80 categories. The evaluation metric is
mean average precision which is widely used in evaluating object detection. The whole
framework is implemented in Caffe\cite{jia2014caffe} Platform.

\subsection{Pascal VOC}
We use Pascal VOC2007 trainval set and Pascal VOC2012 trainval set as our training set,
and VOC2007 test set as testing set, with overall 16k images for training and 5k images
for testing. All models are based on VGG16 architecture since ResNet-101 has limited
gain in this dataset\cite{fu2017dssd}.

We set BPN with two resolutions input(320x320 and 512 x512) and compare them with
the state-of-the-art methods on low, mid and high quality detection scenarios(IoU
thresholds as 0.5, 0.6 and 0.7 respectively). From Table \ref{tab:pascal-voc}, our BPN320
gets 80.3\%, 75.5\% and 66.1\% accuracy in low, mid and high quality detection scenario,
which has already outperformed many detectors(SSD320 and Faster RCNN etc.). Further
we build BPN512 by increasing input size to 512 and BPN512 gets 81.9\%, 77.6\% and
68.3\% in three quality scenarios, which are state-of-the-art results. Our BPN is one-stage
detector, so it can also enjoy the merit for real time inference. BPN320 can make
inference with 32.4fps while BPN512 with 18.9fps on Titan XP GPU cards, which has
significant strength over two-stage detectors. Notably, BPN has very clear advantage in
high quality detection scenario(IoU=0.7).

\begin{table*}[t]
\centering
\caption{Detection results on PASCAL VOC dataset. For VOC 2007, all methods are trained
on VOC 2007 and VOC 2012 {\tt trainval} sets and tested on VOC 2007 {\tt test} set. For some
algorithms there are no public released model, so we emit the results with IoU with 0.7.
Bold fonts indicate the best mAP.}
\footnotesize
\begin{tabular}{c|c|c|c|c|c|c}
\toprule[1.5pt]
\multirow{2}{*}{Method} &\multirow{2}{*}{Backbone} &\multirow{2}{*}{Input size} &\multirow{2}{*}{FPS} &\multicolumn{3}{c}{mAP (\%)} \\
\cline{5-7}
& & &  &IoU@0.5  & IoU@0.6&IoU@0.7 \\
\hline
\textit{two-stage:} & & &  && &\\
Fast R-CNN \cite{girshick2015fast}     &VGG-16 &$\sim1000\times600$  &0.5 &70.0 &62.4&49.4\\
Faster R-CNN \cite{ren2015faster} &VGG-16 &$\sim1000\times600$  &7 &73.2 & 67.7 & 54.4\\
OHEM \cite{shrivastava2016training}      &VGG-16 &$\sim1000\times600$  &7 &74.6  & 68.9 & 55.9\\
HyperNet \cite{kong2016hypernet}        &VGG-16 &$\sim1000\times600$  &0.88 &76.3 & - &-\\
Faster R-CNN \cite{he2016deep} &ResNet-101 &$\sim1000\times600$  &2.4 &76.4  & 69.5 & 57.3\\
ION \cite{bell2016inside}             &VGG-16 &$\sim1000\times600$  &1.25 &76.5 & - &-\\
LocNet \cite{gidaris2016locnet}             &VGG-16 &$\sim1000\times600$  &- &77.5  & - &64.5\\
R-FCN \cite{dai2016r}            &ResNet-101 &$\sim1000\times600$  &9 &80.5  & 73.2 &61.8\\
CoupleNet \cite{zhu2017couplenet}  &ResNet-101 &$\sim1000\times600$  &8.2 &81.7  & 76.6 &66.8\\
\hline
\hline
\textit{one-stage:} & & & & & &\\
YOLO \cite{redmon2016you}       &GoogleNet~\cite{szegedy2015going} &$448\times448$  &45 &63.4 &-&-\\
RON384 \cite{kong2017ron}     &VGG-16 &$384\times384$  &15 &75.4  & 66.8 &54.2\\
SSD300 \cite{liu2016ssd} &VGG-16 &$300\times300$  &46 &77.3  & 72.3 &61.3\\
DSOD300 \cite{shen2017dsod} &DS/64-192-48-1\cite{shen2017dsod} &$300\times300$  &17.4 &77.7  & 73.4 &63.6\\
YOLOv2 \cite{redmon2016yolo9000}   &Darknet-19 &$544\times544$  &40 &78.6  & 69.1 &56.5\\
SSD512 \cite{liu2016ssd} &VGG-16 &$512\times512$  &19 &79.8  & 74.7 &64.0\\
RefineDet320 \cite{zhang2017single}    &VGG-16     &$320\times320$   &40.3 &80.0  & 74.2 &63.6\\
RefineDet512 \cite{zhang2017single}    &VGG-16     &$512\times512$  &24.1 & 81.8  &  76.9& 66.0\\
BPN320(ours)    &VGG-16     &$320\times320$   &32.4 &80.3  & 75.5 &66.1\\
BPN512(ours)   &VGG-16     &$512\times512$  &18.9 &\bf{81.9}  & \bf{77.6} &\bf{68.3}\\
\hline
\bottomrule[1.5pt]
\end{tabular}
\label{tab:pascal-voc}
\end{table*}

\subsection{Ablation Study\label{subsec:abla}}
In this section, we conduct a series of ablation study to analyze the impact of different
components of BPN. We use VOC2007 and VOC2012 {\tt trainval} set as our training set
and test on VOC2007 {\tt test} set. We use mean average precision on three different IoU
thresholds(0.5, 0.6 and 0.7) as our evaluation metric. The results are listed in Table \ref{tab:refinedet}.\\

\noindent{\bf Proposal Quality Improved by CAR:} In this section, we validate the
effectiveness of CAR blocks to improve anchor quality. In Table \ref{tab:matchnum}, we
count the positive anchor numbers per image on different IoU thresholds in original SSD,
FPN and BPN respectively. In original SSD, anchors are generated manually and only a
few anchors matched objects, which is hard to train detectors effectively. In FPN anchors
have been refined by CAR once, and the matched number increases significantly in all
IoU thresholds. Further in BPN where anchors has been refined by CAR twice, more high
quality anchors are generated. Notably, after refined by CAR we have sufficient positive
training samples in high quality levels so that we could conduct gradually increasing
training positive IoU thresholds (0.5, 0.6 and 0.7).
 This experiment shows our CAR blocks can gradually improve anchor qualities and generates more qualified training samples.

 \noindent{\bf Bidirectional Feature Pyramid:} To validate the effectiveness of the
Bidirectional Feature Pyramid, we remove CAR from BPN and compare this
model(dabbed as BPN w / o CAR) with vanilla SSD and SSD w / FPN. Bidirectional Feature
Pyramid is built based on vanilla SSD and all three models are fine-tuned with IoU
threshold as 0.5. In Table \ref{tab:refinedet}, SSD w / FPN outperforms vanilla SSD
because deep semantic features boost feature representations. Further, BPN w / o CAR
outperforms SSD w / FPN in all quality scenarios, which proves the effectiveness of
Bidirectional Feature Pyramid.

\noindent{\bf Level of Cascaded:} In this section, we validate the cascaded level of CAR is
important in training high quality detectors. We list the results in Table
\ref{tab:refinedet}. Firstly, a vanilla SSD model is trained with 0.7 IoU threshold. This
model(Row 2) performs much worse than the baseline(Row 1) trained with 0.5 IoU
threshold in all three quality levels, which validates the fact that insufficient positive
training samples cause overfitting problem. Secondly, we keep only one CAR block of
BPN(dabbed as BPN w / AR), and train this model with 0.5 IoU threshold. The results
show the detection results improves significantly compared with BPN w / o CAR in low
and mid quality scenarios, but not obvious in high quality case(63.6\% vs 63.4\% ). We
further train BPN w / AR with 0.7 IoU threshold and this model(Row 6) also presents
overfitting problem but less severe compared with vanilla SSD. It represents the fact that
anchor refiner can boost detection performance by refine anchor quality but only one
refiner could not directly boost the model. Thirdly, considering results above, we add
two more CAR blocks and joint optimize CAR with different quality settings (0.5,0.5,0.7)
and (0.5,0.6,0.7),  which utilize high quality anchors for training. This two models(Row 7
and Row 8) achieve evident growth especially in high quality scenario(IoU=0.6 and
IoU=0.7, etc.). In conclusion, a single Anchor Refiner is very effective in addressing
overfitting problem in SSD model but to improve the detection performance in high
quality scenarios, Cascaded Anchor Refiner(CAR) is required.

\begin{table*}[t]
\centering
\caption{Detection results on PASCAL VOC dataset. For VOC 2007, all
methods are trained on VOC 2007 and VOC 2012 {\tt trainval} sets and tested on VOC 2007 {\tt test} set.
Original SSD uses six feature maps for prediction, while we use four feature maps to be consistent with BPN,
so the detection result of SSD here is a bit lower. Bold fonts indicate the best mAP.}
\footnotesize \setlength{\tabcolsep}{2.5pt}
\begin{tabular}{c|c|cccc}
\toprule[1.5pt]
\multirow{2}{*}{} &\multirow{2}{*}{Training IoU}&\multirow{2}{*}{mAP@IoU=0.5} &\multirow{2}{*}{mAP@IoU=0.6}& \multirow{2}{*}{mAP@IoU=0.7}\\
  & &  &&&\\
\hline
SSD & 0.5 & 76.3 & 71.0 &60.4  \\
SSD & 0.7 & 68.4 &61.9 &50.8 \\
SSD w / FPN & 0.5 & 77.4 & 72.1 &61.6 \\
BPN w / o CAR & 0.5 & 78.1 & 72.7 &63.4 \\
\hline
BPN w / AR &0.5 &80.0& 74.2 &63.6\\
BPN w / AR &0.7 &78.1 &73.7 &63.1\\
BPN & (0.5, 0.5, 0.7) & 80.0 & 75.1 & 65.4 \\
BPN & (0.5, 0.6, 0.7)&\bf{80.3}&\bf{75.5}&\bf{66.1}\\
\bottomrule[1.5pt]
\end{tabular}

\label{tab:refinedet}
\end{table*}

\subsection{MSCOCO}
In addition to PASCAL VOC, we also evaluate BPN on MSCOCO \cite{lin2014microsoft}.
COCO contains 80 classes objects and about 120k images in {\tt trainval} set. We use {\tt
trainval35k} set for training and test on {\tt test-dev} set.

Table \ref{tab:coco} shows the results on MS COCO test-dev set. BPN320 with VGG-16
achieves 29.6\% AP and when using larger input image size 512, the detection accuracy
of BPN reaches 33.1\%, which is better than all other VGG16-based methods. Notably,
we notice in high quality detection metric $AP_{75}$, BPN is clearly better than other
detectors. Because the objects in COCO dataset are with various scales, so we also
applied multi-scale testing based on BPN320 and BPN512 to reduce the impact of input
size. The improve version BPN320++ and BPN512++ achieves 35.4\% and 37.9\% AP,
which is the state-of-the-art performance of one-stage detectors.

Different from Pascal VOC, deeper backbone such as ResNet can further improve
detection accuracy than VGG16 model. As \cite{zhang2017single} claimed, batch
normalization in ResNet requires at least four images per GPU to obtain precise statistic
information and stable training process. And its deep network structure further enlarges
the memory and computation cost. Thus, we only report the results of VGG16-based
model due to GPU memory and computation limitation.

\begin{table*}[t]\vspace{-0.2in}
\centering
\caption{Detection results on MS COCO {\tt test-dev} set. Bold fonts indicate the best performance.}
\scriptsize
\begin{tabular}{c|c|ccc|cccc}
 \setlength{\tabcolsep}{1.5pt}
Method  &Backbone &AP &AP$_{50}$ &AP$_{75}$ &AP$_{\it S}$ &AP$_{\it M}$ &AP$_{\it L}$\\
\hline
\textit{two-stage:} & & & & & & & & \\
Fast R-CNN \cite{girshick2015fast}  &VGG-16 &19.7 &35.9 &- &- &- &- \\
Faster R-CNN \cite{ren2015faster}  &VGG-16 &21.9 &42.7 &- &- &- &- \\
OHEM \cite{shrivastava2016training}  &VGG-16 &22.6 &42.5 &22.2 &5.0 &23.7 &37.9 \\
ION \cite{bell2016inside}  &VGG-16 &23.6 &43.2 &23.6 &6.4 &24.1 &38.3\\
OHEM++ \cite{shrivastava2016training}  &VGG-16 &25.5 &45.9 &26.1 &7.4 &27.7 &40.3 \\
R-FCN \cite{dai2016r}  &ResNet-101 &29.9 &51.9 &- &10.8 &32.8 &45.0\\
CoupleNet \cite{zhu2017couplenet}  &ResNet-101 &34.4 &54.8 &37.2 &13.4 &38.1 &50.8 \\
Faster R-CNN by G-RMI \cite{huang2017speed}  &Inception-ResNet-v2\cite{szegedy2017inception} &34.7 &55.5 &36.7 &13.5 &38.1 &52.0 \\
Faster R-CNN+++ \cite{he2016deep}  &ResNet-101-C4 &34.9 &55.7 &37.4 &15.6 &38.7 &50.9\\
Faster R-CNN w FPN \cite{lin2016fpn}  &ResNet-101-FPN &36.2 &59.1 &39.0 &18.2 &39.0 &48.2 \\
Faster R-CNN w Cascade RCNN \cite{cai2017cascade}  &VGG16 &26.9 &44.3& 27.8& 8.3& 28.2& 41.1 \\
R-FCN w Cascade RCNN \cite{cai2017cascade}  & ResNet-50 &30.9 &49.9 &32.6 &10.5 &33.1 &46.9    \\
R-FCN w Cascade RCNN \cite{cai2017cascade}  & ResNet-101 &33.3 &52.6 &35.2 &12.1 &36.2 &49.3    \\
\hline
\hline
\textit{one-stage:} & & & & & & & & \\
YOLOv2 \cite{redmon2016yolo9000} &DarkNet-19\cite{redmon2016yolo9000} &21.6 &44.0 &19.2 &5.0 &22.4 &35.5\\
SSD300 \cite{liu2016ssd} &VGG-16 &25.1 &43.1 &25.8 &6.6 &25.9 &41.4\\
RON384++ \cite{kong2017ron} &VGG-16 &27.4 &49.5 &27.1 &- &- &- \\
SSD321 \cite{fu2017dssd}  &ResNet-101 &28.0 &45.4 &29.3 &6.2 &28.3 &49.3\\
DSSD321 \cite{fu2017dssd}  &ResNet-101 &28.0 &46.1 &29.2 &7.4 &28.1 &47.6\\
SSD512 \cite{liu2016ssd}  &VGG-16 &28.8 &48.5 &30.3 &10.9 &31.8 &43.5\\
SSD513 \cite{fu2017dssd}  &ResNet-101 &31.2 &50.4 &33.3 &10.2 &34.5 &49.8 \\
DSSD513 \cite{fu2017dssd}  &ResNet-101 &33.2 &53.3 &35.2 &13.0 &35.4 &51.1 \\
RefineDet320 \cite{zhang2017single}    &VGG-16 &29.4 &49.2 &31.3 &10.0 &32.0 &44.4\\
RefineDet512 \cite{zhang2017single}    &VGG-16 &33.0 &54.5 &35.5 &16.3 &36.3 &44.3 \\
\hline
BPN320    &VGG-16 &29.6 &48.4 &32.3&9.6 &32.5 &44.3  \\
BPN512    &VGG-16  &33.1 &53.1 &36.3 &15.7 &37.0 &44.2 \\
BPN320++    &VGG-16 & 35.4 &55.3 &38.5 &19.0 &37.9 &47.0  \\
BPN512++   &VGG-16 &37.9 &58.0 &41.5 &21.9 &41.1 &48.1  \\
\bottomrule[1.5pt]
\end{tabular}
\label{tab:coco}\vspace{-0.1in}
\end{table*}

\subsection{Quantitative Results and Error Analysis}
\begin{figure}[htp]
\centering
\includegraphics[height=9.5cm]{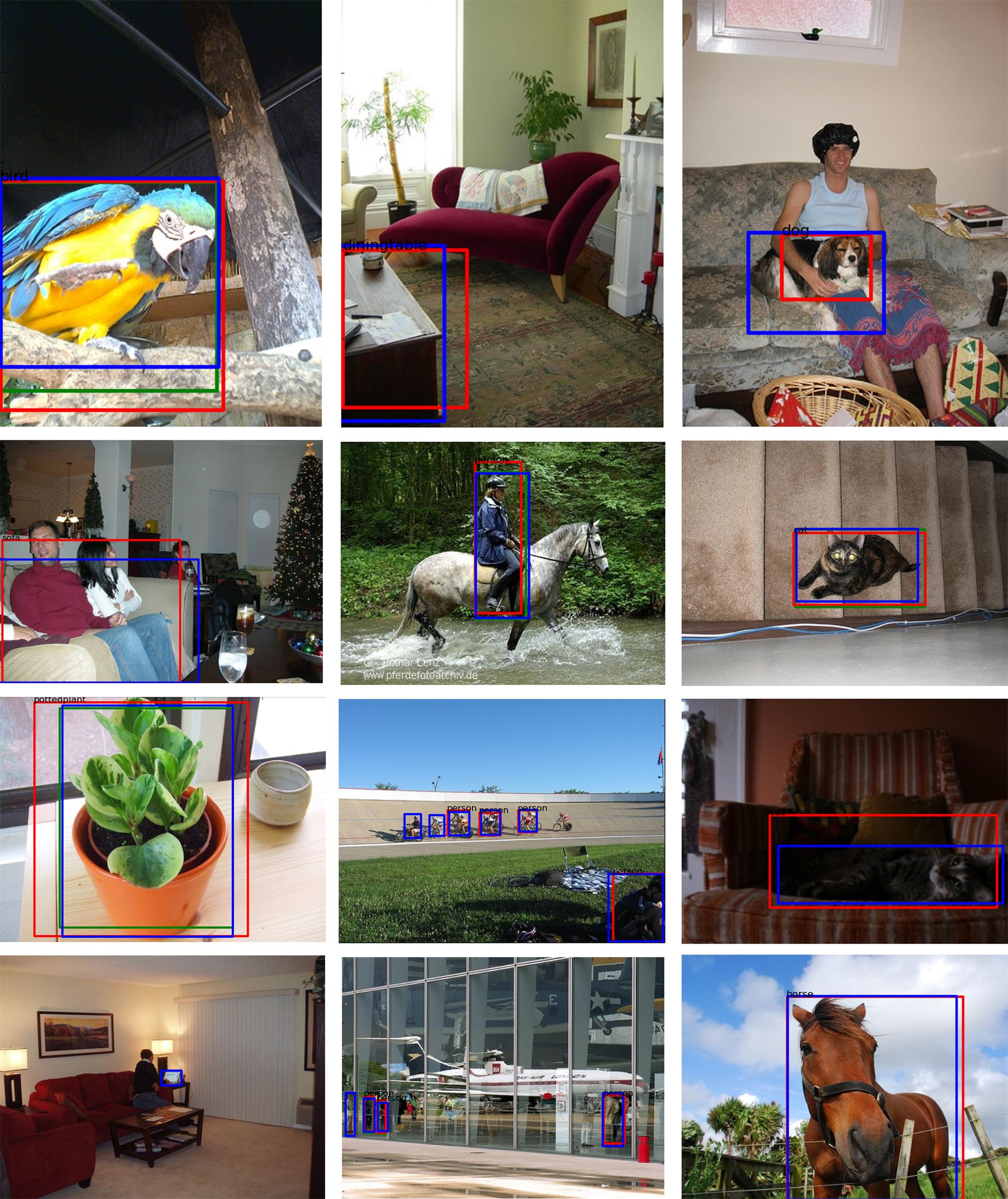}
\caption{Qualitative detection results of our method (blue) on Pascal VOC compared with
other single-shot methods RefineDet \cite{zhang2017single}(red) and SSD \cite{liu2016ssd}(green).
Bounding boxes with confidence scores less than 0.4 are ignored. If there is a missing color, it indicates the corresponding detector fails to detect the object. }
\label{fig:cocoresult}\vspace{-0.05in}
\end{figure}

We show some quantitative results on Pascal VOC in figure \ref{fig:cocoresult}. The
improvement of Cascaded Anchor Refiner and the error analysis are shown in this
section. We compare the results of BPN with other single-stage detector
RefineDet\cite{zhang2017single} and SSD\cite{liu2016ssd}. Results of BPN are illustrated
as blue boxes. Red and Green boxes present the results from
RefineDet\cite{zhang2017single} and SSD\cite{liu2016ssd} respectively. These models
are trained with backbone VGG16 and with the same training setting.  It can be found
that the predictions of BPN match the object boundary more precisely. Also in some hard
cases SSD or RefineDet fail to detect objects while BPN can still work due to stronger
features.

\begin{figure*}[ht]
\centering
\includegraphics[height=8cm]{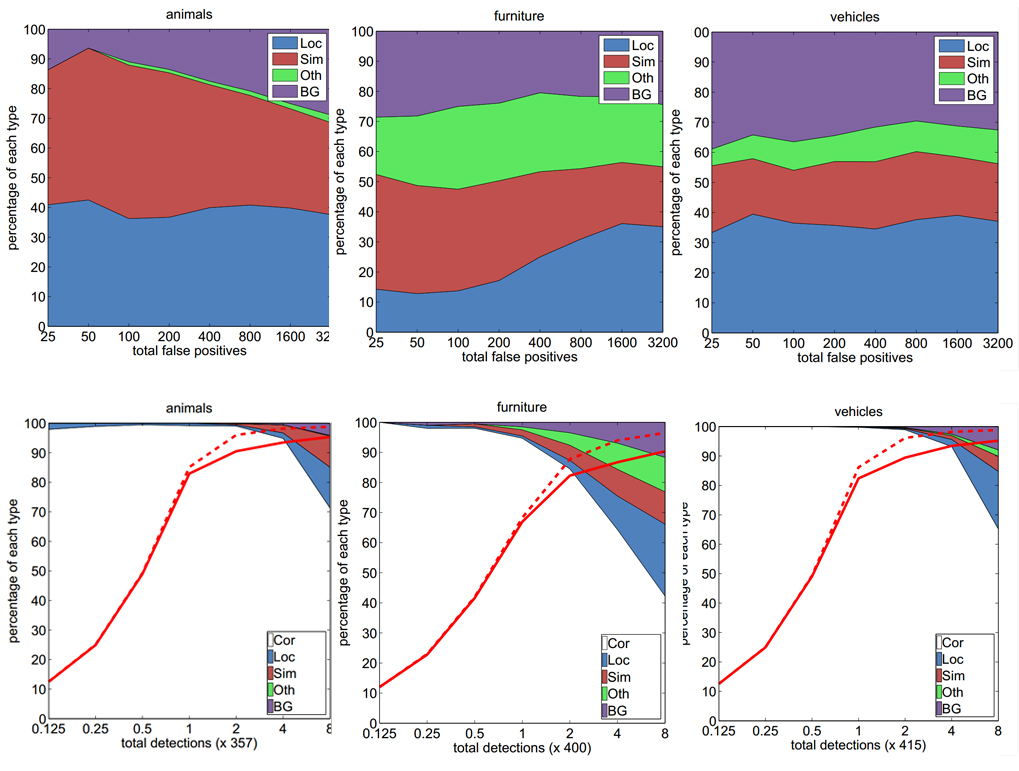}
\caption{Visualization of error analysis for the proposed BPN320 detector on ``animals", ``vehicles", and ``furniture" classes on the VOC 2007 test set. }
\label{fig:error}
\end{figure*}

We also analyze the performance of BPN320 by the detection analysis tool to better
understand the detection result as well as for further improvement. In Figure
\ref{fig:error}, the first row shows the percentage of error type of the top false positive.
The second row shows the fraction of detections that are correct (Cor) or different false
positive types: localization issue(Loc), confusion with similar categories (Sim), with
background (BG) or other errors(Oth). In the first row, BPN320 produces much less
background errors and localization errors compared with vanilla SSD and RefineDet. The
bi-directional feature pyramid strongly boost feature maps and thus is more robust to
complex background information. The cascaded anchor refiners optimize object
locations which leads to more high quality predictions. However, we notice confusion
with similar categories error(Sim) occupies high error ratio in BPN. Similar categories(cow
and cats etc.) make the detector confused and it means the sampling strategy is not
optimal. In this paper, we adopt the same sampling method with previous
work\cite{liu2016ssd} and we argue more effective sampling strategy such as Focal
Loss\cite{lin2017focal} can further improve the detection results. We leave this as future
work. The second row of Figure \ref{fig:error} indicates the majority of BPN's confident
detections are correct.

\section{Conclusions}\vspace{-0.1in}
This paper proposed a novel single-stage detector framework cascaded Bidirectional
Feature Pyramid Networks (BPN) towards high-quality object detection with two major
components: a Bidirectional Feature Pyramid structure for more effective and robust
feature representations and a Cascade Anchor Refinement to gradually refine the quality
of predesigned anchors for more effective training. The proposed method achieves
state-of-the-art results on Pascal VOC and MSCOCO dataset with real-time inference
speed. Future work includes more empirical studies on better backbone networks and
other object detection tasks.

\bibliographystyle{ieee}
\bibliography{egbib}
\end{document}